\documentclass[11pt,a4paper]{article}
\usepackage[hyperref]{acl2019}
\usepackage{times}
\usepackage{latexsym}
\usepackage{url}

\usepackage{amsmath}
\usepackage{booktabs}
\usepackage{graphicx}
\usepackage[utf8]{inputenc}
\usepackage{wrapfig,lipsum}
\usepackage{pgfplots}
\usepackage{pifont}
\usepackage{multirow}
\usepackage{etoolbox, diagbox, colortbl}
\usepackage[group-separator={,}, group-four-digits=true]{siunitx}
\usepackage{verbatim}
\usepackage{mathtools}
\usepackage{chngpage}

\usepackage{amsmath}
\usepackage{amssymb}

\robustify\bfseries
\PassOptionsToPackage{small,bf}{caption}

\aclfinalcopy % Uncomment this line for the final submission

\title{Translating Terminological Expressions in Knowledge Bases \\ with Neural Machine Translation}

\author{Mihael Ar\v{c}an, Daniel Torregrosa and Paul Buitelaar\\ 
 Insight Centre for Data Analytics, Data Science Institute \\
 National University of Ireland Galway \\ 
 {\tt[firstname.lastname]@insight-centre.org}}

\date{}

\begin{document}
\maketitle
\begin{abstract}
Our work presented in this paper focuses on the translation of terminological expressions represented in semantically structured resources, like ontologies or knowledge graphs. The challenge of translating ontology labels or terminological expressions documented in knowledge bases lies in the highly specific vocabulary and the lack of contextual information, which can guide a machine translation system to translate ambiguous words into the targeted domain. Due to these challenges, we evaluate the translation quality of domain-specific expressions in the medical and financial domain with statistical as well as with neural machine translation methods and experiment domain adaptation of the translation models with terminological expressions only. Furthermore, we perform experiments on the injection of external terminological expressions into the translation systems. Through these experiments, we observed a significant advantage in domain adaptation for the domain-specific resource in the medical and financial domain and the benefit of subword models over word-based neural machine translation models for terminology translation. %Nevertheless, through the specific and unique terminological expressions, subword segmentation within NMT does not outperform a word based neural translation model.
\end{abstract}

\section{Introduction}

Most of the labels stored in semantically structured resources, like ontologies, taxonomies or knowledge graphs, are documented in English only \cite{gracia2012challenges}. Applications in information retrieval, question answering or knowledge management that use these monolingual resources are therefore limited to the language in which the information, namely terms, labels or metadata, is stored. To enable knowledge access across languages, these resources need to be enriched with multilingual information. This enhancement can enable information extraction on documents beyond English, e.g. for cross-lingual business intelligence in the financial domain \cite{ORiainCBDKT13,MTS2013-Arcan}, providing information related to an ontology label, e.g. \textit{other intangible assets}, in Spanish, German or Italian.

%Since manual translation of semantically structured resources is a time consuming and expensive process, since it is mostly performed by domain experts, we engage and evaluate the performance of statistical machine translation (SMT) and neural machine translation (NMT) methods to automatically translate the domain-specific expressions. 
%We evaluate both statistical machine translation (SMT) and neural machine translation (NMT) systems for automatic translation of the domain-specific expressions, motivated by the invasive manual procedure domain experts would alternatively endure. 

Due to the large success of neural machine translation (NMT)  \cite{kalchbrenner13emnlp,Sutskever:2014:SSL:2969033.2969173}, we compare the use of NMT and statistical machine translation (SMT) \cite{brown93:tmo} when translating terminological expressions in isolation, i.e. when they do not form part of a whole sentence. % evaluate its translation performance against the usage of statistical machine translation (SMT) \cite{brown93:tmo}, comparing which method handles better the translation of terminological expressions in isolation, i.e. without any contextual information.
This is motivated by the invasive manual procedure domain experts would alternatively endure. %Within the neural architecture, the encoder transforms a terminological expression into a vector representation, which is passed to the decoder generating the target expression word by word from the vector representation.
Although automatically generated translations of these domain-specific expressions are far from perfect, studies have shown significant productivity gains when human translators are supported by machine translation output rather than starting a translation task from scratch \cite{federico2012measuring,laubliassessing,green2013efficacy}.

For both translation methods, we translated the ontology labels in the medical and financial domain, documented in the International Classification of Diseases (ICD) and in the International Financial Reporting System (IFRS) ontology. Furthermore, we translated the Wikipedia titles, which represent a mixture of generic as well as domain-specific expressions. Since large parallel in-domain corpora are not always available, the vocabulary of these specific resources, which usually document only a few thousand terms, is translated with translation models trained on generic parallel data only. Furthermore, we evaluate how terminological expressions, within the same domain, can contribute to the adaptation of translation models to improve the translation quality. Due to the fact that the terminological expressions are usually highly domain-specific and appear infrequent in parallel corpora, we compare the availability of injecting external knowledge into the translation process to address the out-of-vocabulary (OOV) issue. Since the information in ontologies or knowledge graphs is often defined only in one language, mostly in English, we translate these expressions from English to German. 

Our work shows that while using models trained (and tuned) on generic data, subword NMT models demonstrate a better performance than SMT on the task of domain-specific expression translation. Furthermore, our experiments on domain adaptation with terminological expressions showed a significant improvement of translation quality with the usage of NMT for all targeted resources. Although the results for external knowledge injection show better performance for the SMT approach on the open-domain Wikipedia evaluation set, NMT outperforms the SMT approach when translating the domain-specific ontology labels in the medical and financial domains.

\section{Related Work}
Most of the previous work on translation knowledge resources, e.g. ontologies or taxonomies, tackled this problem by accessing multilingual lexical resources, such as EuroWordNet or IATE \cite{Declerck06,Cimiano:2010:NOL:1834624.1834625}. Their work focuses on the identification of the lexical overlap between the ontology labels and the multilingual resource. Since the replacement of the source and target vocabulary guarantees a high precision but a low recall, external translation services, such as BabelFish, SDL FreeTranslation tool or Google Translate, were used to overcome this issue \cite{FuBO09,Espinoza:2009}. Additionally, ontology label disambiguation was performed by \newcite{Espinoza:2009} and \newcite{Mccrae:11}, where the structure of the ontology along with existing multilingual ontologies was used to annotate the labels with their semantic senses. Furthermore, \newcite{mccrae2016domain} show positive effects of different domain adaptation techniques, i.e., using Web resources as additional bilingual knowledge, re-scoring translations with Explicit Semantic Analysis (ESA) and language model adaptation for automatic ontology translation. A different approach on ontology label disambiguation was shown in \newcite{arcan2015:ACL}, where the authors identified relevant in-domain parallel sentences and used them to train an ontology-specific SMT system.

%As a lexical resource, Wikipedia with its rich semantic knowledge was used as a resource for bilingual term identification in the context of SMT. \newcite{Tyers:2008} extracts bilingual dictionary entries from Wikipedia to support the machine translation system. Based on exact string matching they query Wikipedia with a list of around 10,000 noun lemmas to generate the bilingual dictionary. 
A different approach is to use rich semantic knowledge bases such as Wikipedia for bilingual term identification in the context of MT. \newcite{Tyers:2008} extract bilingual dictionary entries by looking for exact string matches in Wikipedia. Besides the Wikipedia interlanguage links, \newcite{Erdmann:2009} enhance their bilingual dictionary by using redirection page titles and anchor text within Wikipedia. To cast the problem of ambiguous Wikipedia titles, \newcite{Niehues:11} and \newcite{Arcan:14} use the information of Wikipedia categories and the text of the articles to provide the SMT system with domain-specific bilingual knowledge. %This research showed that using the lexical information stored in this knowledge base improves the translation of highly domain-specific vocabulary. 

Due to the recency of NMT, terminology translation with neural models is still less examined. This can be explained due to the issue that neural models are incapable of translating rare words, like domain-specific expressions, because they have a fixed size of vocabulary. Nevertheless, without the help of subword segmentation, \newcite{DBLP:journals/corr/LuongSLVZ14} utilized the out-of-vocabulary issue by a post-processing step that replaced every unknown word with the usage of a dictionary. Differently to the post-processing step, \newcite{chatterjee-EtAl:2017:WMT1} propose a mechanism that guides an existing NMT decoder with the ability to prioritize and adequately handle translation candidates provided by the external resource. In the case of domain adaptation, most work focuses on \textit{fine tuning}, where an out-of-domain model is further trained on in-domain data \cite{DBLP:conf/acl/SennrichHB16,luong2015stanford,DBLP:journals/corr/ServanCS16}. In addition to the \textit{fine-tuning} method, \newcite{P17-2061} tune the neural model with in- and out-of-domain data, whereby they use tags to annotate the domains within the used corpora. Differently to these approaches, which focused on document translation, our research focuses entirely on the translation of short and domain-specific expression documented in knowledge bases, without any contextual information guiding the adaptation or translation approach.

\section{Methodology}

We use the approaches to SMT and NMT to translate the terminological expressions, with a special focus on the performance of NMT and how it handles the translation of expressions infrequently found in the parallel training data. Therefore, we explore how the translation quality may benefit from the usage of subword segmentation, which helps to overcome the issue of vocabulary limitation in the neural network approach. We furthermore perform domain adaptation with domain-specific expressions and experiment with different approaches and methods on injecting external knowledge into the translation process. These methods are detailed in the following subsections.

%\subsection{Neural Machine Translation}
\subsection{Neural Machine Translation}

In this work, we use the RNN architecture, consisting of an encoder and a decoder. The encoder reads an input sequence $x=(x_1,...,x_n)$ and the decoder predicts a target sequence $y=(y_1,...,y_n)$. 
The encoder and decoder interact via a soft-attention mechanism \cite{DBLP:journals/corr/BahdanauCB14,DBLP:journals/corr/LuongPM15}, which comprises of one or multiple attention layers.

\begin{equation} \label{eq:rnn-enc-hidden}
h^{l}_{i} = h^{l-1}_{i} + f_{rnn}(h^{l-1}_{i}, h^{l}_{i-1})
\end{equation} 

$h^{l}_{i}$ corresponds to the hidden state at step $i$ of layer $l$. $h^{l}_{i-1}$ represents the hidden state at the previous step of layer $l$ while $h^{l-1}_{i}$ means the hidden state at $i$ of $l-1$ layer. $E \in \mathbb{R}^{m\times K_x}$ is a word embedding matrix, $W \in \mathbb{R}^{n\times m}$, $U \in \mathbb{R}^{n\times n}$ are weight matrices, with $m$ being the word embedding size and $n$ the number of hidden units. $K_x$ is the vocabulary size of the source language. Thus, $E_{x_{i}}$ refers to the embedding of $x_{i}$, and $e_{pos,i}$ indicates the positional embedding at position $i$. 
% %\vspace{-2mm}
% %\paragraph{RNN-based NMT.}
% %  \label{ssub:rnn_based_nmt}
% In \ac{RNN} models, networks change as new inputs (previous hidden state and the token in the line) come in, and each state is directly connected to the previous state only. Therefore, the path length of any two tokens with a distance of $n$ in RNNs is exactly $n$. Its architecture enables adding more layers, whereby two adjoining layers are usually connected with residual connections in deeper configurations. Equation \ref{eq:rnn-enc-hidden} displays $h^{l}_{i}$, where 
% %accordingly the $l$th encoder layer by which 
% $f_{rnn}$ is usually a function based on \ac{GRU}
% %~\cite{cho2014learning} 
% or \ac{LSTM}
% %~\cite{hochreiter1997long}
% . The first layer is then represented as $h^{0}_{i} = f_{rnn}(WE_{x_{i}}, U h^{0}_{i-1})$. Additionally, the initial state of the decoder is commonly initialized with the average of the hidden states or the last hidden state of the encoder. 

\subsection{Domain Adaptation with Terminological Expressions}
In this work, we experimented how both translation models, either SMT and NMT, can be adapted to the targeted domain and the text style, namely the translation length, word order or compounding in the German language, of the domain-specific terms. To adapt the log-linear weights of the SMT system to the resource type and domain, we rerun MERT \cite{Och:2003:MER:1075096.1075117} using a minimal development set of the available domain-specific resources (Table~\ref{tab:eval_dom_adapt}). %Therefore, we start the re-tuning approach with in-domain data from the already adapted weights of the different models, based on the generic development set (Table~\ref{tab_eval_dom_adapt}).
%To evaluate, if pre-tuned weights have any influence on domain adaptation, we applied two strategies: first we start the re-tuning approach from the already adapted weights of the different models, based on the generic development set (\textit{generic$\rightarrow$dom.-specific} in Table~\ref{tab_eval_dom_adapt}). %The second strategy starts adapting the weights from the unadapted, uniformed weights (\textit{dom.-specific only}). 
For the domain adaptation within NMT, we used the models trained on the generic data (or generic and Wikipedia resource) and retrained the weights in the neural network again only on a minimal development set ($\approx$ 1,000 terms for ICD/IFRS) of each domain.
We additionally perform weight adaptation across different domains, to evaluate, if the properties of domain-specific expressions of the development set can improve the translation quality regardless of the domain.

\subsection{Integration of Terminological Knowledge into Machine Translation}
\label{subsec:term_inject}

Due to the fact that domain-specific bilingual information might be missing and cannot be learned from the parallel sentences, some of the terminological expressions may not be automatically translatable with an SMT or NMT system. Therefore, we use the information obtained from Wikipedia or use the ontology labels as in-domain knowledge ontology\footnote{We used the development sets as external, in-domain knowledge, which was not used in the training process for this experiment.} to improve the translation of expressions, which are not known to the translation systems.
 
%\paragraph{XML Markup } 
%With the help of the \textbf{XML markup approach within SMT}, 

We guide the SMT system with external knowledge that can be directly passed to the decoder by specifying the translation of a specific span of the source sentence. In the case of multiple translations of the same source span, a score can be used to indicate the level of association between the source and target phrases. In the case of using Wikipedia titles as external knowledge, we perform two experiments. In the first experiment, we set all probabilities of the translation candidates to 1.0. In the second experiment, we calculate the cosine similarity (Equation \ref{eq:cos}) between the knowledge base vocabulary \textit{\pmb x} and the vocabulary of the Wikipedia abstracts \textit{\pmb y} associated with the titles, which form our translation candidates. 

\begin{equation}
cos( x, y) = \frac { x \cdot y}{|| x|| \cdot || y||}
\label{eq:cos}
\end{equation}

As an example for the targeted medical domain, we obtain a higher cosine similarity for the preferred candidate \textit{orbit--orbita}\footnote{the socket in the skull which contains the eye} than for the \textit{orbit--Umlaufbahn}\footnote{the gravitationally curved path of one object around a point or another body} translation candidate in the technical domain.

\begin{adjustwidth}{0.25in}{0.25in}
    \textit{ $<$n translation="orbita$\mid\mid$umlaufbahn" prob="0.872 $\mid\mid$ 0.512"$>$orbit$<$/n$>$ }
\end{adjustwidth}
\vspace{2mm}
 
Furthermore, we compared three different methods for injecting terminological expressions into the SMT framework, i.e., \texttt{exclusive}, \texttt{inclusive} and \texttt{constraint}. In the \texttt{exclusive} setting, only the proposed translations are used for the input phrase. Translation candidates stored in the phrase table and overlapping with that span are ignored. Differently, the proposed translations compete with the translation candidates in the phrase table, if the \texttt{inclusive} setting is selected. In the \texttt{constraint} setting, the proposed translations compete with phrase table choices that contain the specified translation. 

%\vspace{-3mm}
%\paragraph{Providing External Knowledge to NMT}
The NMT tool used in this work, i.e. OpenNMT, allows providing \textbf{external knowledge for replacement} of unknown words in the neural models. %\footnote{OpenNMT option \texttt{-phrase\_table}} 
Instead of inserting the \texttt{\textless unk\textgreater} token for an unknown word, it will lookup in the external knowledge for a possible translation. In addition to providing external knowledge, OpenNMT enables to substitute unknown words with source words that have the highest attention weight. %\footnote{OpenNMT option \texttt{-replace\_unk}} 
We use both options to evaluate the performance of the translation quality.

\section{Experimental Setting}
In this section, we give an overview on the datasets and the translation tools used in our experiment. Furthermore, we give insight into the SMT and NMT evaluation techniques, considering the English to German translation direction.

\subsection{Training Datasets}
For a broader domain coverage of the generic training dataset necessary for the SMT system, we merged parts of JRC-Acquis 3.0 %\footnote{\url{https://ec.europa.eu/jrc/en/language-technologies/jrc-acquis}} 
\cite{Steinberger:2006}, Europarl v7 %\footnote{\url{http://www.statmt.org/europarl/}}
\cite{Koehn:2005} and OpenSubtitles2013 \cite{Tiedemann:2012}, obtaining a training corpus of almost two million sentences, containing around 38M running words (Table~\ref{tab:stats_general}).\footnote{For reproducibility and future evaluation we take the first one-third part of each corpus. All datasets and translation models will be published at the time the paper is published.} %This generic SMT system is used as a %general baseline, which we compare against the domain-specific models generated with different sentence selection methods. Furthermore we use the generic SMT system in combination with the smaller domain-specific models to evaluate different approaches when combining generic and domain-specific data together.

Due to the challenging task on terminology translation, we perform an additional experiment where we combine the generic corpus of two million sentences with Wikipedia titles ($\sim$876k), which have an interlanguage link between the English and German language in Wikipedia.

\begin{table}[t]
\setlength{\tabcolsep}{2pt}
\centering
%\small
\begin{tabular}{lr|S[table-format=9.0]S[table-format=9.0]}
\toprule
& & {English} & {German} \\
\midrule
Generic  & Lines & \multicolumn{2}{S[table-format=9.0]}{1924646}\\
dataset & Words & 37947852 & 35728314 \\
(training) & Vocab. & 237108 & 493448 \\
\midrule
Wikipedia & Entries & \multicolumn{2}{S[table-format=9.0]}{876657}\\
(knowledge & Words & 2468864 & 2364991 \\
injection) & Vocab. & 394835 & 445333 \\
\midrule
Generic & Lines & \multicolumn{2}{S[table-format=9.0]}{2762053}\\
+Wikipedia & Words & 39841507 & 37562721 \\
(training)& Vocab. & 564519 & 860239 \\
\midrule
\midrule
ICD  & Lines & \multicolumn{2}{S[table-format=9.0]}{1000}\\
ontology & Words & 6028 & 5928 \\
(develop.) & Vocab. & 1380 & 1584 \\
\midrule
IFRS  & Lines & \multicolumn{2}{S[table-format=9.0]}{1000}\\
ontology & Words & 10288 & 10844 \\
(develop.)& Vocab. & 741 & 1089 \\
\midrule
Wikipedia & Lines & \multicolumn{2}{S[table-format=9.0]}{50121}\\
(develop.) & Words & 183866 & 170496 \\
& Vocab. & 43614 & 50972 \\
\midrule
\midrule
ICD  & Lines & \multicolumn{2}{S[table-format=9.0]}{915}\\
ontology & Words & 5763 & 5742 \\
(evaluation) & Vocab. & 1257 & 1492 \\
\midrule
IFRS  & Lines & \multicolumn{2}{S[table-format=9.0]}{1000}\\
ontology & Words & 10049 & 10533 \\
(evaluation)& Vocab. & 733 & 1088 \\
\midrule
Wikipedia & Lines & \multicolumn{2}{S[table-format=9.0]}{49861}\\
(evaluation) & Words & 171442 & 151575 \\
& Vocab. & 45336 & 52126 \\
\bottomrule
\end{tabular}
%\vspace{-2mm}
\caption{Statistics for the bilingual training and evaluation datasets (develop\textit{-}ment / vocab\textit{-}ulary).} % (\textit{Vocabulary} denotes the number of unique words in the data set).}
\label{tab:stats_general}
\end{table}

\subsection{Evaluation Datasets}
%For our evaluation tasks we apply SMT and NMT translation models on domain-specific expressions. %For a general overview and comparison of the SMT and NMT performance, we use the \textbf{WMT15}\footnote{Workshop on Statistical Machine Translation, \\ \url{http://www.statmt.org/wmt15/index.html}} English-German evaluation set. For the main task on domain-specific expression translation, 
For the task on domain-specific expression translation, we evaluate the translation quality based on the specific  ontology labels in the medical and in the financial domain. In addition to that, we perform translations on Wikipedia titles. Due to the size of the entire Wikipedia, we extend the evaluation set to approximately 50,000 entries.  %Since the titles can be domain-specific as well as generic, we extend the evaluation set to approximately 50,000 entries. 
On the other hand, the smaller datasets of the ontological resources are defined by the availability of the alignment across languages or the resources itself, which are due to their domain specificity rather small.

%\paragraph{ICD ontology} 
For our experiments we used the \textbf{ICD-10 ontology}\footnote{\url{http://www.who.int/classifications/icd/en}} as the gold standard. The ICD ontology, translated into \num{43} languages, is used to monitor diseases and to report the general health situation of the population in a country. This stored information supports providing an overview of the national mortality rate and appearance of diseases of countries inside World Health Organisation. %organisation.

%\paragraph{IFRS ontology} 
The \textbf{IFRS ontology} (International Accounting Standards Board) \cite{RePEc:wbk:wbpubs:2288} is used for providing electronic financial reports for auditing. The terms contained within this taxonomy are frequently long (on average \num{11} tokens) and are entirely composed of noun phrases.

%\paragraph{Wikipedia Titles} 
\textbf{Wikipedia}\footnote{\url{http://www.wikipedia.org}} is a multilingual, freely available encyclopaedia that was built by a collaborative effort of voluntary contributors. All combined Wikipedias hold approximately \num{40} million articles with more than \num{27} billion words in more than \num{293} languages, making it the largest collection of freely available knowledge. %\footnote{\url{http://en.wikipedia.org/wiki/Wikipedia:Size_comparison}} 
%In our work we use the DBpedia \cite{dbpedia-swj} repository (version 2016-04), which provides structured knowledge of Wikipedia. %The DBpedia project aims to extract structured content from the knowledge added to the Wikipedia repository. %DBpedia allows users to semantically query relationships and properties of Wikipedia resources, including links to other related datasets.%

%With the heavily interlinked information base, Wikipedia forms a rich lexical and semantic resource. Besides a large number of articles, it also holds a hierarchy of categories that Wikipedia articles are tagged with. It includes knowledge about named entities, domain-specific terms and word senses. 

\subsection{Machine Translation tools}

%\paragraph{Moses}
For our SMT translation task, we use the statistical translation toolkit \textbf{Moses} \cite{Koehn:2007}, where the word alignments were built using the GIZA++ toolkit \cite{Och:2003}. The KenLM toolkit \cite{Heafield-kenlm} was used to build a 5-gram language model. 
%\paragraph{OpenNMT}

For the NMT task, we use \textbf{OpenNMT}~\cite{2017opennmt}, a generic deep learning framework mainly specialized in sequence-to-sequence (seq2seq) models covering a variety of tasks such as machine translation, summarisation, image to text, and speech recognition. Due to computational complexity, the vocabulary in NMT models had to be limited. In order to overcome this limitation, we used byte pair encoding (BPE) to generate subword units \cite{journals/corr/SennrichHB15}. BPE is a form of data compression that iteratively replaces the most frequent pair of bytes in a sequence with a single, unused byte. We used the default OpenNMT parameters, i.e. \num{2} layers, \num{500} hidden LSTM units, input feeding enabled, batch size of \num{64}, \num{0.3} dropout probability and a dynamic learning rate decay. We trained the networks for \num{13} epochs and report the results in Section \ref{sec:eval}.

\subsection{Evaluation Metrics}
The automatic translation evaluation is based on the correspondence between the machine translation hypothesis and reference translation (gold standard). For the automatic evaluation we used the BLEU~\cite{Papineni:2002}, METEOR \cite{denkowski:lavie:meteor-wmt:2014} and chrF3~\cite{popovic:2015:WMT} metrics.

\textbf{BLEU} (Bilingual Evaluation Understudy) is calculated for individual translated segments (n-grams) by comparing them with a dataset of reference translations. Those scores, between \num{0} and \num{100} (perfect overlap), are then averaged over the whole \textit{evaluation dataset} to reach an estimate of the translation's overall quality. 
\textbf{METEOR} (Metric for Evaluation of Translation with Explicit ORdering) is based on the harmonic mean of precision and recall, whereby recall is weighted higher than precision. Along with exact word (or phrase) matching it has additional features, i.e. stemming, paraphrasing and synonym matching. In contrast to BLEU, the metric produces good correlation with human judgement at the sentence or segment level. 
\textbf{chrF3} is a character n-gram metric, which showed very good correlations with human judgements on the WMT2015 shared metric task~\cite{stanojevic-EtAl:2015:WMT}, especially when translating from English into morphologically rich(er) languages.

\section{Evaluation}
\label{sec:eval}

In this section, %we report a general quality comparison between SMT and NMT translation models based on the WMT15 dataset. Furthermore 
we evaluate the translation quality of ontology labels and Wikipedia titles with the SMT and NMT methods. Additionally, we explore the performance of domain adapted systems when translating in- and out-of-domain knowledge. In the final experiment, we inject in-domain lexical knowledge into the translation systems.

\subsection{Translation Evaluation of Terminological Expressions}

In this evaluation section, we focus on the domain-specific vocabulary documented in the ICD and IFRS ontology, as well as the Wikipedia titles. The domain-specific entries in these datasets appear infrequent in the training data and unlike the training dataset, where the entries appear in the context of a sentence, the test set only contains one entry, i.e. term, per line.

% and are translated in isolation, i.e. without any contextual information. %, which may help in the disambiguation approach of ambiguous expressions.

%Similarly to the WMT15 evaluation experiment, 
We observed that concatenating Wikipedia titles with the generic corpus does not always improve the translation quality when translating the ontology labels (Table~\ref{tab:eval_dom_adapt}). %%%OLD TABLE REF
Focusing on the SMT method, the performance in terms of BLEU improves for the ICD ontology labels (6.39 to 7.40), whereby the performance of the IFRS ontology labels drops (10.51 vs 9.03). Differently to that, the vocabulary similarity of the merged generic and Wikipedia dataset helps significantly to improvement the translation quality of the Wikipedia titles.

When we applied NMT models, we observed that the subword model (BPE32k) always outperforms the word-based model. This can be explained by the fact that the word-based model is limited to a vocabulary of 50,000 words, and cannot learn the representations of terminological expressions infrequently appearing in the training data. The subword model overcomes this limitation by segmenting these expressions into smaller units.
Comparing the best NMT performance (BPE32k) with SMT, the former generates better translations for the IFRS labels and for the Wikipedia entries, when only the generic dataset is used. Conversely, the SMT approach  generates better translations for the medical terms documented in the ICD ontology.

\begin{table*}[]
    \centering
    %\small
    \begin{tabular}{ccccccccc}
        \toprule
         & \multicolumn{4}{c}{Generic Dataset} & \multicolumn{4}{c}{Generic+Wikipedia} \\
         \midrule
         & \multicolumn{2}{c}{English} & \multicolumn{2}{c}{German} & \multicolumn{2}{c}{English} & \multicolumn{2}{c}{German} \\
         \textbf{Words} & inCorpus & OOV & inCorpus & OOV & inCorpus & OOV & inCorpus & OOV \\
         \cmidrule{2-9}
        ICD & 1,020 & 235 & 928 & 561 & 1,125 & 120 & 1,055 & 434 \\
        IFRS & 705 & 28 & 926 & 162 & 706 & 27 & 941 & 147\\
        Wikipedia & 21,174 & 22,672 & 22,261 & 28,659 & 34,136 & 9,710 & 36,982 & 13,938\\
        \midrule
       \textbf{Terms}  & inCorpus & OOV & inCorpus & OOV & inCorpus & OOV & inCorpus & OOV \\
          \cmidrule{2-9}
        ICD & 47 & 868 & 32 & 883 & 64 & 851 & 50 & 865\\
        IFRS & 31 & 969 & 30 & 970 & 31 & 969 & 30 & 970  \\
        Wikipedia & 3,389 & 46,472 & 3,732 & 46,129 & 6,240 & 43,621 & 6,777 & 43,084\\
        \bottomrule
    \end{tabular}
    %\vspace{-2mm}
    \caption{Vocabulary overlap between the ICD, IFRS and Wikipedia titles and the training data on the word and term level.}
    \label{tab:found_oov}
\end{table*}

To better understand the automatic evaluation, we studied the vocabulary overlap of the ICD, IFRS labels and Wikipedia titles and the training datasets (Table \ref{tab:found_oov}). We observed a better overlap between the training dataset and the IFRS ontology in comparison to the ICD ontology. This shows that the translations of financial labels in the IFRS ontology can be learned better in comparison to the medical labels in the ICD ontology. We further learned that the ratio of known and unknown words between Wikipedia titles and the generic dataset is close to similar. This can be explained due to the appearance of named entities in Wikipedia, such as \textit{kothamangalam}\footnote{Kothamangalam is a municipality in the eastern part of Ernakulam district in the Indian state of Kerala} or \textit{claddagh}.\footnote{Claddagh is an area close to the centre of Galway} Although concatenating Wikipedia titles with the training set significantly reduced the number of OOV words, this does not entirely reflect the automatic evaluation metric scores in Section \ref{tab:eval_dom_adapt}. When studying the appearance of entire terms in the training dataset, the number of observed terms significantly drops. On average, between three and six percent of the ICD, IFRS and Wikipedia expressions appear in the training dataset. This percentage slightly increases when Wikipedia titles are added to the training data.

\begin{table}
\setlength{\tabcolsep}{2pt}
\centering
%\small
\begin{tabular}{lcc|cc}
\toprule
& \multicolumn{2}{c}{Generic Dataset} & \multicolumn{2}{c}{Generic+Wikipedia} \\
\cmidrule{2-5}
& English  & German & English & German \\
\midrule
ICD & 67.93 & 37.60 & 65.95 & 35.92\\
IFRS & 94.81 & 70.86 & 93.58 & 68.10\\
Wikipedia & 29.75 & 17.01 & 38.86 & 24.89\\
\bottomrule
\end{tabular}
%\vspace{-2mm}
\caption{Vocabulary coverage (in \%) between the ICD, IFRS and Wikipedia titles and the neural translation model vocabulary.}
\label{tab:coverage}
\end{table}

Furthermore, we calculated the vocabulary coverage of the neural models regarding the evaluation dataset. As shown in Table \ref{tab:coverage}, we observed a high coverage of the financial vocabulary documented in the IFRS ontology within the neural model vocabulary. Similarly to the observations in Table \ref{tab:found_oov}, Wikipedia has the lowest vocabulary coverage within the neural models. Interestingly, the coverage drops slightly for ICD and IFRS if the generic dataset is concatenated with the Wikipedia entries, since some generic expressions within the Wikipedia entries exclude domain-specific expressions occurring in the generic dataset.

These results support the automatic evaluation (Table \ref{tab:eval_dom_adapt}), %%%OLD TABLE REF
where the IFRS labels show better translation quality in terms of the evaluation metrics in comparison to the ICD labels or Wikipedia titles. When concatenating the generic dataset with the Wikipedia entries, the translation quality of the Wikipedia evaluation set improves significantly in comparison to translations generated by the generic system only.

A manual evaluation showed the drawbacks of terminology translation with the word-based NMT model are a result of the vocabulary limitations of the neural architecture. For example, the expression \textit{bacterial} within the medical term \textit{other bacterial diseases} could not be translated (\textit{sonstige} \texttt{<unk>} \textit{krankheiten}) with the word-based neural model, since the word seldom appears in the training data, hence it was not included into the vocabulary. Instead, SMT and subword models translated the term correctly (\textit{sonstige bakterielle krankheiten}).
%other bacterial diseases sonstige bakterielle krankheiten andere bakteriellen krankheiten  sonstige <unk> krankheiten andere bakterielle krankheiten
%other functional intestinal disorders sonstige funktionelle darmstörungen andere funktionalen hi störungen  andere funktionale <unk> sonstige funktionelle darmstörungen
Further advantages of the NMT models are shown within the example \textit{injury of blood vessels at hip and thigh level}, where on the one hand, the SMT approach translates the medical expressions literally word by word, and, on the other hand, does not translate the ambiguous word \textit{vessel}\footnote{meaning of \textit{boat, ship} or meaning of \textit{container})} correctly as \textit{blut\textbf{gefäßen}} in the medical domain. The trained subword models translate the entire term correctly, translating multi-word expressions into a German compound (\textit{blood vessels} $\rightarrow$ \textit{blutgefäßen}), as well as reorder the terms correctly on the target side, where \textit{höhe} (en. \textit{level}) was moved from the end of the expression to the beginning, i.e. (\textit{höhe der hüfte und des oberschenkels}). Similarly, a better handling of German compounds can be observed for the financial term \textit{aggregate adjustment}, which was translated word by word by the SMT approach, i.e. \textit{kumulierte anpassung}, but was correctly provided as a compound with the subword neural models, i.e. \textit{gesamtanpassung}. For the same term, the word-based NMT model omitted the translation of the word \textit{aggregate} and provided the German word \textit{anpassung} (en. \textit{adjustment}). Although we learned that subword models significantly outperform all other approaches, we also observed wrong translations. Due to the word segmentation, \textit{heartburn} was segmented into \textit{heart} and \textit{burn}. Because of this, both segments were translated segment by segment and generated a literal translation as a compound in German, i.e. \textit{herzverbrennung}.\footnote{\textit{heart} $\rightarrow$ \textit{herz} and \textit{burn} $\rightarrow$ \textit{verbrennung}}

\subsection{Domain Adaptation with Terminological Expressions}

\begin{table*}[ht!]
%\small
\centering
\setlength{\tabcolsep}{1pt}
\sisetup{detect-weight=true,detect-inline-weight=math,table-format=2.2,round-precision=2,round-integer-to-decimal,round-mode=places}
\begin{tabular}{lr|SSS|SSS|SSS}
\toprule 
&  & \multicolumn{3}{c|}{BLEU} &  \multicolumn{3}{c|}{METEOR} &  \multicolumn{3}{c}{chrF3} \\
\midrule[\heavyrulewidth] 
\multirow{12}{*}{\rotatebox[origin=c]{90}{\parbox[c]{6cm}{\centering Generic Dataset}}} & SMT models & {ICD$_{eval}$} & {IFRS$_{eval}$} & {Wiki$_{eval}$} & {ICD$_{eval}$} & {IFRS$_{eval}$} & {Wiki$_{eval}$} & {ICD$_{eval}$} & {IFRS$_{eval}$} & {Wiki$_{eval}$}\\
\cmidrule{2-11} 
& Baseline & 6.39 & 10.51 & 12.49 & 14.11 & 16.02 & 21.81 & 41.89 & 45.03 & 49.14 \\
& {ICD$_{dev}$}  & \bfseries 8.02 & 12.14 & 13.07 & \bfseries 14.76 & 16.82 & 21.76 & 42.17 & 45.06 & 49.16 \\
& {IFRS$_{dev}$}  & 7.17 & \bfseries 14.24 & 13.14 & 13.83 & \bfseries 17.77 & 22.01 & 41.64 & \bfseries 47.7 & \bfseries 49.63 \\
& {Wiki$_{dev}$}  & 6.12 & 8.58 & \bfseries 14.47 & 13.27 & 15.02 &\bfseries 22.1 & \bfseries 42.66 & 43.81 &  49.58 \\
\arrayrulecolor{white}\midrule\arrayrulecolor{black}   
& NMT models & {ICD$_{eval}$} & {IFRS$_{eval}$} & {Wiki$_{eval}$} & {ICD$_{eval}$} & {IFRS$_{eval}$} & {Wiki$_{eval}$} & {ICD$_{eval}$} & {IFRS$_{eval}$} & {Wiki$_{eval}$}\\
\cmidrule{2-11} 
& Baseline & 3.20 & 9.35 & 8.2 & 8.65 & 13.18 & 10.13 & 20.46 & 28.35 & 21.38\\
& {ICD$_{dev}$}  & \bfseries 20.89 & 3.55 & 3.21 & \bfseries 31.06 & 12.14 & 6.86 & \bfseries 37 & 15.08 & 11.06\\
& {IFRS$_{dev}$}  & 2.53 & \bfseries 58.17 & 7.39 & 13.62 & \bfseries 65.48 & 14.71 & 20.9 & \bfseries 65.47 & 18.05\\
& {Wiki$_{dev}$}  & 1 & 0.74 & \bfseries 26.27 & 3.99 & 5.27 & \bfseries 27.08 & 7.51 & 8.32 & \bfseries 26.64\\
\arrayrulecolor{white}\midrule\arrayrulecolor{black}   
&  NMT\textsubscript{BPE} models & {ICD$_{eval}$} & {IFRS$_{eval}$} & {Wiki$_{eval}$} & {ICD$_{eval}$} & {IFRS$_{eval}$} & {Wiki$_{eval}$} & {ICD$_{eval}$} & {IFRS$_{eval}$} & {Wiki$_{eval}$}\\
\cmidrule{2-11} 
& Baseline & 4.29 & 13.55 & 13.51 & 19.22 & 30.31 & 26.9 & 38.99 & 43.48 & 45.01 \\
& {ICD$_{dev}$}  & \bfseries 50.15 & 5.76 & 9.41 & \bfseries 58.68 & 19.05 & 20.71 & \bfseries 72.82 & 35.05 & 38.29 \\
& {IFRS$_{dev}$}  & 4.86 & \bfseries 75.03 & 10.78 & 20.83 & \bfseries 81.48 & 23.9 & 40.7 & \bfseries 88.22 & 42.35 \\
& {Wiki$_{dev}$}  & 1.53 & 1.52 & \bfseries 41.19 & 7.69 & 8.98 & \bfseries 42.34 & 22.47 & 19.35 & \bfseries 55.76\\
\midrule 
\multirow{12}{*}{\rotatebox[origin=c]{90}{\parbox[c]{6cm}{\centering Generic + {Wiki}pedia}}} & SMT models & {ICD$_{eval}$} & {IFRS$_{eval}$} & {Wiki$_{eval}$} & {ICD$_{eval}$} & {IFRS$_{eval}$} & {Wiki$_{eval}$} & {ICD$_{eval}$} & {IFRS$_{eval}$} & {Wiki$_{eval}$}\\
\cmidrule{2-11} 
& Baseline & 7.40 & 9.03 & 37.81 & 13.41 & 14.97 & 31.03 & 39.9 & 42.12 & 62.44 \\
& {ICD$_{dev}$}  & \bfseries 9.11 & 11.88 & 31.72 & \bfseries 15.37 & 16.19 & 29.23 & \bfseries 43 & 44.83 & 59.89\\
& {IFRS$_{dev}$}  & 8.06 & \bfseries 14.21 & 31.98 & 14.12 & \bfseries 17.97 & 29.58 & 41.64 & \bfseries 48.13 & 60.59\\
& {Wiki$_{dev}$}  & 4.36 & 5.33 & \bfseries 37.98 & 11.27 & 11.89 & \bfseries 30.41 & 36.69 & 35.8 & \bfseries 61.36\\
\arrayrulecolor{white}\midrule\arrayrulecolor{black}   
& NMT models & {ICD$_{eval}$} & {IFRS$_{eval}$} & {Wiki$_{eval}$} & {ICD$_{eval}$} & {IFRS$_{eval}$} & {Wiki$_{eval}$} & {ICD$_{eval}$} & {IFRS$_{eval}$} & {Wiki$_{eval}$}\\
\cmidrule{2-11} 
& Baseline & 2.26 & 9.25 & 22.87 & 6.25 & 12.84 & 20.28 & 15.39 & 27.55 & \bfseries 40.09 \\
& {ICD$_{dev}$}  &  \bfseries 19.53 & 0 & 0 & \bfseries 28.68 & 4.38 & 0 & \bfseries 34.42 & 8.63 & 4.11\\
& {IFRS$_{dev}$}  & 0 & \bfseries 54.23 & 0.05 & 5.95 & \bfseries 60.54 & 2.32 & 13.46 & \bfseries 60.75 & 7.97\\
& {Wiki$_{dev}$}  & 0.31 & 0 & \bfseries 23.23 & 1.66 & 2.19 & \bfseries 23.9 & 3.55 & 3.34 & 23.19\\
\arrayrulecolor{white}\midrule\arrayrulecolor{black}   
&  NMT\textsubscript{BPE} models & {ICD$_{eval}$} & {IFRS$_{eval}$} & {Wiki$_{eval}$} & {ICD$_{eval}$} & {IFRS$_{eval}$} & {Wiki$_{eval}$} & {ICD$_{eval}$} & {IFRS$_{eval}$} & {Wiki$_{eval}$} \\
\cmidrule{2-11} 
& Baseline & 3.58 & 11.82 & 35.49 & 15.45 & 27.2 & 42.83 & 33.5 & 38.92 & 53.02\\
& {ICD$_{dev}$}  &  \bfseries 47.43 & 0 & 23.13 & \bfseries 56.76 & 4.38 & 30.71 & \bfseries 70.97 & 8.63 & 48.49\\
& {IFRS$_{dev}$}  & 5.17 & \bfseries 69.72 & 0.05 & 19.59 & \bfseries 78.14 & 2.32 & 38.35 & \bfseries 86.01 & 7.97\\
& {Wiki$_{dev}$}  & 2.28 & 4.59 & \bfseries 44.28 & 10.49 & 17.51 & \bfseries 47.02 & 26.47 & 28.28 & \bfseries 59.11\\
\bottomrule 

\end{tabular}
%\vspace{-2mm}
\caption{Evaluation on domain adaptation based on different terminological datasets.}
\label{tab:eval_dom_adapt}
\end{table*}

We performed the experiment on domain adaptation using the development sets of the resources mentioned before, which adapts the weights of the log-linear models in SMT or the weights in the neural network architecture.

As seen in Table~\ref{tab:eval_dom_adapt}, using the in-domain development set of the same domain, the translation quality improves when compared with the SMT system, trained and tuned on generic data. For example, the BLEU score of translations of the ICD labels increases from \num{6.39} to \num{8.02}, if the weights are adapted to the targeted domain. Similarly, significant improvements are observed for the IFRS and Wikipedia evaluation dataset.

When using the models with generic sentences and Wikipedia knowledge, the translation quality improves for the ICD (\num{9.11} vs \num{7.40}) and IFRS ontology labels (\num{14.21} vs. \num{9.25}). On the other side, the experiment shows minor improvements for the Wikipedia titles (\num{37.98} vs. \num{37.81}). %The translation quality of the was harmed with the added Wikipedia knowledge, which was also already observed in generic evaluation approach (Table~\ref{tab:auto_eval}).

%A comparison between domain adaptation with pre-adapted weights and adapted weights on terminological resources only did not show any significant changes for the ICD and Wikipedia dataset. Differently, the translation performance of the IFRS ontology labels drops, if the weight adaptation was done only with the in-domain development set. Additionally we observed that MERT converged much faster in the setting, if weights were previously adapted on the generic development set. 

In comparison to the domain adaptation of the SMT method, domain adaptation with NMT methods demonstrates an improvement for the domain-specific ontologies, ICD and IFRS. The BLEU score for the word-based model improves from  \num{8.02} (SMT adapted model) to  \num{20.89}  for the ICD ontology labels, and from  \num{14.24}  to  \num{58.17}  for the IFRS financial labels. Similarly, the BLEU score for the Wikipedia evaluation set increases from  \num{14.47}  to  \num{26.27} .

\begin{table}[ht!]
%\small
\centering
\setlength{\tabcolsep}{10pt}
\begin{tabular}{lcc}
\toprule
& English & German \\
\cmidrule{2-3}
ICD & 58.31 & 38.47\\
IFRS & 74.76 & 49.81\\
Wikipedia & 32.19 & 25.55 \\
\bottomrule
\end{tabular}
%\vspace{-2mm}
\caption{Vocabulary overlap (in \%) between the ICD, IFRS and Wikipedia evaluation dataset and the development dataset used for domain adaptation.}
\label{tab:eval_dev}
\end{table}

Table \ref{tab:eval_dev} illustrates the vocabulary overlap between the evaluation set and the development set used for domain adaptation. The  translation improvements relate to the overlapping coverage, supporting the IFRS improvement from around  \num{10 } BLEU points to around  \num{55 } using the generic neural model and  \num{75 } BLEU points when the Generic+Wikipedia dataset is used for domain adaptation.

In general, the subword NMT models further improve the translation quality for all evaluation sets, which is due to the fact that, on the one hand, subword models overcome the OOV issue of word-based neural models, and, on the other hand, adjust the weights of the entire network accordingly to the targeted domain. Although the word based neural model adjust the network in the same way, they face the issue of unknown words, due to the vocabulary limitations. Similar improvements were shown in ~\newcite{W17-4713}, where the adaptation approaches on document translation outperformed the  original  generic  NMT  system as well as a strong phrase-based SMT system.

\begin{table*}
\centering %\small
%\tiny
\setlength{\tabcolsep}{3pt}
\sisetup{detect-weight=true,detect-inline-weight=math,table-format=2.2,round-precision=2,round-integer-to-decimal,round-mode=places}
\begin{tabular}{lll|SSS|SSS|SSS}
\toprule 
& & & \multicolumn{3}{c|}{BLEU} &  \multicolumn{3}{c|}{METEOR} &  \multicolumn{3}{c}{chrF3} \\ 
\midrule 
\multirow{11}{*}{\rotatebox[origin=c]{90}{\parbox[c]{5cm}{\centering SMT}}} &  \multicolumn{2}{l|}{ICD}  & {inclus.} & {constr.} & {exclus.} & {inclus.} & {constr.} & {exclus.} & {inclus.} & {constr.} & {exclus.} \\
\cmidrule{2-12}
&  \multicolumn{2}{l|}{Wiki $p=1$}  & 4.91 & 3.56 & 3.29 & 14.27 & 12.01 & 12.01 & 42.66 & 38.97 & 38.99 \\
&  \multicolumn{2}{l|}{Wiki $p=cos(x,y)$}  &  \bfseries 5.03  & 3.42 & 3.37 & \bfseries 14.58 & 12.04 & 12.18 & \bfseries 43.44 & 38.99 & 39.17 \\
&  \multicolumn{2}{l|}{In-dom $p=1$}  & 7.1 &  \bfseries 8.05   &  \bfseries 8.05   & 14.23 & \bfseries 14.59 & \bfseries 14.59 & 42.61 & \bfseries  43.07 & \bfseries  43.07\\
\arrayrulecolor{white}\midrule\arrayrulecolor{black}  
&  \multicolumn{2}{l|}{IFRS}   & {inclus.} & {constr.} & {exclus.} & {inclus.} & {constr.} & {exclus.} & {inclus.} & {constr.} & {exclus.} \\
\cmidrule{2-12}
&  \multicolumn{2}{l|}{Wiki $p=1$} & 10.23 & 6.54 & 6.44 & 15.64 & 12.4 & 12.29 & 45.09 & 38.35 & 38.31 \\
&  \multicolumn{2}{l|}{Wiki $p=cos(x,y)$} &  \bfseries 10.54   & 6.52 & 6.42 & \bfseries 16.17 & 12.39 & 12.26 & \bfseries 46.13 & 38.45 & 38.49 \\
&  \multicolumn{2}{l|}{In-dom $p=1$} & 26.72 & 29.65 &  \bfseries 29.69   & 22.47 & 24.3 & \bfseries 24.31 & 54.44 & \bfseries 57.3 &  \bfseries 57.30\\
\arrayrulecolor{white}\midrule\arrayrulecolor{black} 
&   \multicolumn{2}{l|}{Wiki}   & {inclus.} & {constr.} & {exclus.} & {inclus.} & {constr.} & {exclus.} & {inclus.} & {constr.} & {exclus.} \\
\cmidrule{2-12}
&  \multicolumn{2}{l|}{Wiki $p=1$}  & 16.59 & 17.03 & 17.11 & 23.46 & 23.48 & 23.52 & 52.22 & 51.62 & 51.62 \\
&  \multicolumn{2}{l|}{Wiki $p=cos(x,y)$}  & 34.7 & 42.68 &  \bfseries 42.84   & 39.49 & 44.91 & \bfseries 44.95 & 74.27 & 80.62 & \bfseries 80.64 \\
& In-dom $p=1$ &  &  /  &  /  &  /  &  /  &  /  &  /  &  / &  / &  /\\
\midrule
\midrule
\multirow{4}{*}{\rotatebox[origin=c]{90}{\parbox[c]{3cm}{\centering NMT}}} &  &  & {ICD} & {IFRS}  & {Wiki}  & {ICD}  & {IFRS}  & {Wiki}  & {ICD}  & {IFRS}  & {Wiki} \\
\midrule 
& \multirow{2}{*}{Wikipedia} & {1-gram} &  \bfseries 6.88   & 11.71 &  \bfseries 15.74   & \bfseries 12.94 & 15.55 & \bfseries 22.16 & 35.2 & \bfseries 36.49 & \bfseries 46.86\\
&  & {Lex-align} & 6.46 &  \bfseries 11.90  & 13.52 & 12.92 & \bfseries 15.66 & 20.23 & \bfseries 36.49 & 36.32 & 44.52\\
\cmidrule{2-12}
& \multirow{2}{*}{In-domain} & {1-gram} &  \bfseries 10.41  &  \bfseries 14.66   &  /  & 16.69 & \bfseries 18.2 &  /  & 43.05 & \bfseries 44.35 &  / \\
&  & {Lex-align} & 10.39 & 14.65 &  /  & \bfseries 16.72 & 18.19 &  /  & \bfseries 43.09 & 44.34 &  /\\
\bottomrule
\end{tabular}
%\vspace{-2mm}
\caption{Evaluation of term injection into SMT and NMT system (term injection methodology: inclus. = inclusive, constr. = constraint, exclus. =  exclusive). }
\label{tab:term_inject}
\end{table*}

Although the SMT system was adapted on short terminological expressions or ontology labels, the generated translations in German often begin with an determiner, i.e. \textit{die, der} or \textit{das} (en. \textit{the}). Although this is not an issue when translating terminological expressions in documents, this behaviour is penalised by the evaluation metrics when the hypotheses are compared with the evaluation set, which does not list determiners at the beginning of the terminological expressions. As an example, domain-specific expressions \textit{disclosure whether loans \dots}, \textit{financial effect of changes \dots} or \textit{adjustments for increase \dots} were always generated  with a German determiner in the beginning, e.g. \textit{\textbf{die} offenlegung \dots}, \textit{\textbf{die} auswirkung} or \textit{\textbf{die} anpassungen für erhöhen \dots}, respectively. This behaviour was not observed with the subword neural models, which provided translations without determiners, e.g. \textit{offenlegung , ob darlehen \dots}, \textit{finanzielle auswirkungen von änderungen \dots} or \textit{anpassungen bei der erhöhung}. Furthermore, the neural models were better adapted to the particular domain in comparison to the SMT adapted models. As an example, the financial term \textit{equity} (within \textit{issue of equity}) and the medical term \textit{orbita} (within \textit{disorders of orbit}) were wrongly translated into their most dominant meaning, i.e. \textit{andere rücklagen frage der \textbf{gerechtigkeit}}\footnote{en. \textit{justice}} and \textit{störungen der \textbf{umlaufbahn}},\footnote{in the meaning of \textit{satellite orbit}} whereby the subword neural models provided correct translations, i.e. \textit{ausgabe von \textbf{eigenkapital}, andere rücklagen} for the financial and \textit{störungen der \textbf{orbita}} for the medical domain.

\subsection{External Knowledge Injection into Machine Translation}

\begin{table*}[t]
%\small
\centering
\setlength{\tabcolsep}{8pt}
\begin{tabular}{cccccc}
\toprule
& \multicolumn{2}{c}{Vocabulary} & \multicolumn{3}{c}{Entire Terms} \\
\cmidrule{2-6}
& English & German & English Term & German Term  & English+German Terms  \\
\midrule
ICD & 80.27 & 37.46 & 4.59 & 3.38 & 2.51\\
IFRS & 77.89 & 28.21 & 0.40 & 0.20 & 0.10 \\
WIkipedia & 68.28 & 55.86 & 0.09 & 0.11 & 0.00 \\
\bottomrule
\end{tabular}
\vspace{-2mm}
\caption{Vocabulary overlap (in \%) between the ICD, IFRS and Wikipedia titles and the external knowledge injection with Wikipedia entries.}
\label{tab:extknow_overlap}
\end{table*}

In the final experiment, we compared the performance when injecting external knowledge to the SMT and NMT systems. To simulate the common scenario, we only use the models trained with generic data only and inject in-domain ontology labels\footnote{For the knowledge injection, we used the labels from the development set, which was not used for adjusting the translation models in this experiment.} and the Wikipedia titles as an external knowledge into the translation process. 

As mentioned in Section \ref{subsec:term_inject}, we give a probability of \num{1} to all translation candidates. In the second setting, we adapt the probability depending on the cosine similarity between the vocabulary of the ontology and the vocabulary of the Wikipedia abstracts associated with the Wikipedia titles. We learned from Table~\ref{tab:term_inject} that the adapted probability (\textit{Wiki vocab. / adapted prob.}) shows minor improvements over the non-adapted probabilities. This demonstrates that adding all Wikipedia entries as an external resource does not improve the translation performance. A similar observation was observed in \newcite{srivastava2017improving}, where the authors also used Wikipedia entries with a similar outcome. In detail, the performance drops from  \num{6.39}  to  \num{5.03}  for the ICD ontology labels, with similar results for the IFRS ontology (\num{10.54}  vs  \num{10.51}). Focusing on the Wikipedia evaluation dataset, the similar vocabulary helps to outperform the generic model without the external knowledge (\num{42.84}  vs  \num{12.49}). %\footnote{We did not perform any ICD or IFRS domain vocabulary injection for the Wikipedia evaluation set, due to small sizes of the ICD, IFRS resources.}
Additionally, we used the development set of each ontology as an external resource and injected the in-domain translation candidates into the translation process of the generic model. Compared to the usage of Wikipedia as the external resource, we observe an increase of  \num{3} BLEU points for the ICD dataset (\num{8.05}), and almost \num{20} for the IFRS dataset (\num{29.69}).
Focusing on Wikipedia knowledge injection into a generic neural translation model, the unigram replacement method (\textit{unigram rep.} in Table~\ref{tab:term_inject}) shows best performance for the ICD and Wikipedia evaluation set, whereby the lexical alignment (\textit{lex. alignment}), where the provided word is chosen based on the closest source word embedding vector, shows best results for the IFRS ontology labels. 
When using in-domain knowledge as an external resource, the translation quality of the ontology labels improves over the Wikipedia injected knowledge, with almost identical performance between unigram replacement and lexical alignment. Compared to the SMT performance on injecting an external resource, we gained translation improvement for the ICD ontology labels (\num{10.41}  vs.  \num{8.05}) but observed a significant drop in terms of BLEU for the IFRS ontology labels (\num{14.66}  vs.  \num{29.69}).

Due to the minimal translation improvement while adding Wikipedia entries as external knowledge, we explored the vocabulary overlap between the evaluation sets and the vocabulary of the injected knowledge (Table \ref{tab:extknow_overlap}). Although there is a larger vocabulary overlap between the resources in English and German language independently, only a small overlap exists if the entire term, either in English or in German, is considered. Finally, when considering the English and German terms as a single unit, only  \num{23}  terms documented in the ICD ontology were identified in the injected external knowledge, e.g. \textit{dystonia - dystonie}, \textit{peritonsillar abscess - peritonsillarabszess} or \textit{cerebral palsy - infantile zerebralparese}. For the IFRS ontology, only one term pair was identified, i.e. \textit{depreciation - abschreibung}. This outcome showed the differences between the precise and highly-specific vocabulary and translation pairs in the medical and financial domain compared to the the Wikipedia entries, which are written collaboratively by largely anonymous volunteers.

\section{Conclusion}

This work presents a performance comparison between SMT and NMT when translating highly domain-specific expressions, such as medical or financial terminological expressions, as documented in the ICD and IFRS ontologies. Furthermore, we performed experiments on translating Wikipedia titles, which can be domain-specific as well as generic expressions. We showed that the Wikipedia resource can be beneficial in the translation approach, but due to the lexical ambiguity of the Wikipedia titles, the translation candidates have to be ranked or filtered accordingly to the targeted domain. We also demonstrated that domain adaptation with only terminological expressions significantly improves the translation quality, which is specifically evident if an existing generic neural network is retrained with a limited vocabulary of the targeted domain. Our future work further focuses on quality assurance of the domain-specific expressions and the injection of multi-word terminological expressions into the NMT system to improve the translation of domain-specific vocabulary stored in semantically structured resources.

\section*{Acknowledgments}
This publication has emanated from research conducted with the financial support of Science Foundation Ireland (SFI) under Grant Number SFI/12/RC/2289 (Insight).

%\linespread{1}

\bibliography{emnlp2018.bib}
\bibliographystyle{acl_natbib}

\end{document}